\documentclass[conference]{IEEEtran}
\IEEEoverridecommandlockouts
\usepackage{cite}
\usepackage{amsmath,amssymb,amsfonts}
\usepackage{algorithmic}
\usepackage{graphicx}
\usepackage{textcomp}
\usepackage{xcolor}
\usepackage{multirow}%
\usepackage{amsthm}%
\usepackage{mathrsfs}%
\usepackage{textcomp}%
\usepackage{booktabs}%
\usepackage{listings}%
\usepackage{xfrac}
\usepackage{siunitx}
\usepackage{tabularx}
\usepackage{float}
\usepackage{gensymb}

\def\BibTeX{{\rm B\kern-.05em{\sc i\kern-.025em b}\kern-.08em
    T\kern-.1667em\lower.7ex\hbox{E}\kern-.125emX}}
\begin{document}

\title{Integrated Digital Reconstruction of Welded Components: Supporting Improved Fatigue Life Prediction\\
\thanks{The Innovation Fund Denmark supported this study, project: CeJacket - Create Disruptive solutions for Design and Certification of Offshore Wind Jacket foundations, project number 6154-00017B. Ørsted A/S supported the Wenglor MLWL131 scanner used for this system.}
}

\author{\IEEEauthorblockN{1\textsuperscript{st} Anders F. Mikkelstrup}
\IEEEauthorblockA{\textit{Department of Materials and Production} \\
\textit{Aalborg University}\\
Aalborg, Denmark \\
Email: afm@mp.aau.dk}
\and
\IEEEauthorblockN{2\textsuperscript{nd} Morten Kristiansen}
\IEEEauthorblockA{\textit{Department of Materials and Production} \\
\textit{Aalborg University}\\
Aalborg, Denmark \\
Email: morten@mp.aau.dk}
}

\maketitle

\begin{abstract}
In the design of offshore jacket foundations, fatigue life is crucial. Post-weld treatment has been proposed to enhance the fatigue performance of welded joints, where particularly high-frequency mechanical impact (HFMI) treatment has been shown to improve fatigue performance significantly. Automated HFMI treatment has improved quality assurance and can lead to cost-effective design when combined with accurate fatigue life prediction. However, the finite element method (FEM), commonly used for predicting fatigue life in complex or multi-axial joints, relies on a basic CAD depiction of the weld, failing to consider the actual weld geometry and defects. Including the actual weld geometry in the FE model improves fatigue life prediction and possible crack location prediction but requires a digital reconstruction of the weld. Current digital reconstruction methods are time-consuming or require specialised scanning equipment and potential component relocation. The proposed framework instead uses an industrial manipulator combined with a line scanner to integrate digital reconstruction as part of the automated HFMI treatment setup. This approach applies standard image processing, simple filtering techniques, and non-linear optimisation for aligning and merging overlapping scans. A screened Poisson surface reconstruction finalises the 3D model to create a meshed surface. The outcome is a generic, cost-effective, flexible, and rapid method that enables generic digital reconstruction of welded parts, aiding in component design, overall quality assurance, and documentation of the HFMI treatment.
\end{abstract}


\section{Introduction}
When designing jacket foundations for the offshore industry, fatigue life is commonly a determining factor. Several methods for post-weld treatment have been proposed, significantly improving the fatigue performance of welded joints. One of which is high-frequency mechanical impact (HFMI) treatment, which has been proven effective throughout the literature, outperforming conventional methods such as burr-grinding \cite{Yildirim2013}. HFMI treatment essentially works by locally hammering the weld toe at a frequency exceeding 90 Hz. By smoothing the transition between weld and base material, the stress concentration is reduced, while the plastic deformation of the weld toe introduces compressive residual stress; hence, counteracting the tensile stresses that cause fatigue. 

However, HFMI treatment must adhere to strict guidelines from the International Institute of Welding (IIW) to yield the expected fatigue life improvements. The IIW guidelines \cite{MarquisGaryB2016} include geometrical requirements for the resulting treatment groove, such as width, depth and placement of the groove, which can be difficult for a human operator to adhere to when operating a heavy and vibrating tool for extended periods. Additionally, operator bias cannot be omitted, which can further affect the resulting fatigue performance of the treated weld \cite{Alden2020}. 

As a result, certification agencies like DNV \cite{DNV} do not recommend using HFMI treatment at the design stage. To overcome this issue, \cite{Mikkelstrup2022} proposed a methodology for performing automated HFMI treatment to improve quality assurance. The proposed methodology used an industrial manipulator with input from a 3D scanner to identify the areas requiring treatment. The study showed a significant improvement in the treatment variance evaluated based on the quantitative quality metrics proposed by IIW, i.e. groove depth, width and placement. However, the proposed method for quality assurance only evaluates the geometry and is thus unable to provide any fatigue life prediction. Hence, accurate fatigue life prediction can support a more cost-effective design through material reduction, as the expected fatigue improvement can be documented. 

In the case of complex or multi-axially loaded joints, the finite element method (FEM) is commonly applied for fatigue life prediction. However, typically FEM modelling is based on a simple and generic CAD representation of the weld, which does not consider the actual weld geometry and weld defects, which are well-known to influence the stress concentrations \cite{Larsen2021}. To improve the FE model, the actual weld geometry can be included in the model, i.e. by digitally reconstructing the welded geometry using 3D scanning. This method has been shown to improve the fatigue life prediction and the prediction of the possible crack location \cite{Larsen2021}\cite{Spath2023}.

However, scanning the welded joints for digital reconstruction is often manual and time-consuming or requires specialised scanning tools and possible relocation of the part, making it unfit for use in a production setup \cite{Rodriguez-Gonzalvez2017}. Furthermore, the resulting point cloud should have the highest possible resolution and accuracy, preferably below 0.1 \si{mm}, to ensure that the areas of potential stress concentrations are captured in the reconstructed geometry. Therefore, handheld techniques are unsuitable, as these commonly offer a maximum achievable spatial resolution of up to 0.2 \si{mm}, even when using external targets \cite{Rodriguez-Gonzalvez2017}. 

Therefore, this work proposes a generic framework for cost-effective, flexible, and rapid digital reconstruction of welded components for enhanced FEM modelling. The overall aim is to support improved fatigue life prediction for a more cost-effective design of welded components.
The proposed framework is developed to be applied as an integrated part of the setup in which automated post-weld treatment is performed, such as the setup in \cite{Mikkelstrup2022}. The suggested approach relies only on an industrial manipulator with a line scanner to comprehensively scan the sample from multiple angles to produce a complete surface representation. Prevalent image processing techniques and simple filtering techniques combined with non-linear optimisation are applied for aligning and merging the overlapping point clouds. This is followed by a screened Poisson surface reconstruction to create the full watertight 3D representation suitable for FEM modelling. Note that the proposed framework is not limited to setups for post-weld treatment but could be used in a setup for automated welding. 

The paper is organised as follows: Section \ref{sec:method} presents the processing steps and the main principles behind the framework. The experimental setup for validating the framework is presented in Section \ref{sec:exp}, while the results are presented in Section \ref{sec:res}. Lastly, the concluding remarks and suggestions for future works are presented in Section \ref{sec:conc}.

\section{Method}\label{sec:method}
\begin{figure}[!t]
	\centering
	\includegraphics[width=1\linewidth]{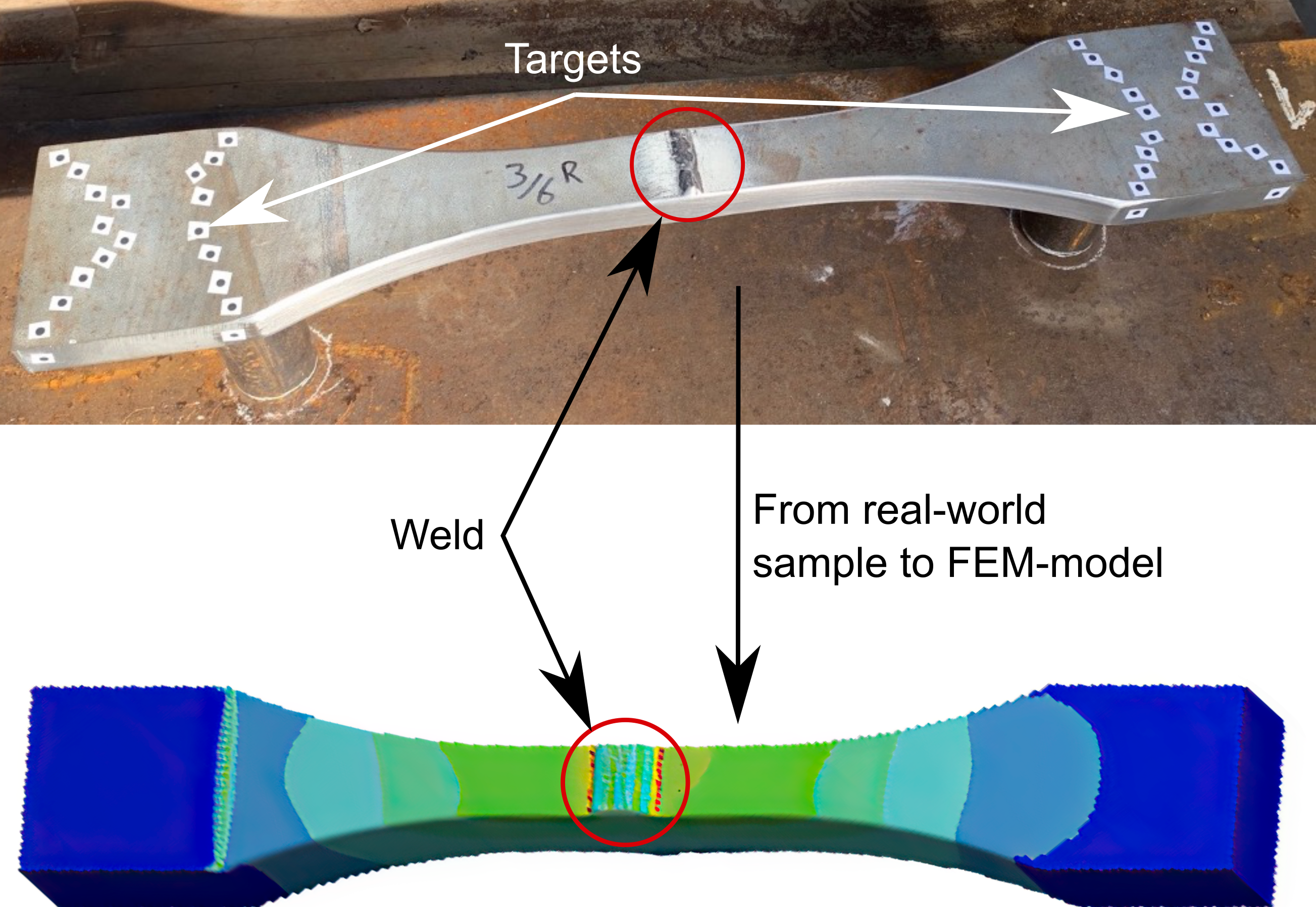}
	\caption{The figure illustrates the transformation of a real-world welded dog bone sample into an FE model, used to predict fatigue life and potential crack location. The depicted targets aid in aligning the point clouds to form a comprehensive surface representation of the sample. Note that the red areas of the FEM model indicate the highest stresses and, consequently, the most probable locations for cracks.}
	\label{fig:sample}
\end{figure}

The proposed framework is based on small-scale S355 dog bone samples of welded joints for the offshore industry, as shown in Fig.~\ref{fig:sample}. However, the framework can be applied to any component suitable and feasible to scan. A complete and accurate representation of the entire sample is required to achieve the most accurate fatigue life prediction, including any misalignment or distortion. Although, a high resolution around the area of the weld is essential. Any contours excessively smoothed in the reconstruction will significantly affect the predicted fatigue life. Therefore, it is necessary to scan the sample from a multitude of orientations and positions, resulting in multiple point clouds that must be aligned, merged, and meshed to be suitable for use in FE modelling.

Estimating spatial transformations that align the point clouds are known as point cloud registration. The point cloud alignment commonly consists of two stages: a rough initial alignment and a secondary fine-tuning \cite{MinetolaPaolo2012Tioa}. The rough initial alignment is usually performed based on the scanner's position during acquisition, whereas the fine-tuning of the spatial transformation can roughly be divided into targetless registration and target-based registration. 

The most well-known algorithm for targetless point cloud registration is perhaps the ICP (iterative closest point) proposed by \cite{BeslPJ1992Amfr}. However, as these methods commonly rely on matching unique point pairs in the point clouds, they are not well suited for texture-less and plain surfaces and suffer from a lack of robustness.

Instead, the target-based approach uses artificial targets that have been well distributed and placed in a unique pattern on the object, such that several targets are visible from all scanning angles. Identifying and locating the target pairs in the point clouds makes it possible to compute the spatial transformation between them. In this work, the target-based approach has been chosen, as it is beneficial when scanning texture-less objects that do not have distinctive features to be used for aligning the point clouds, and results in an accurate and robust registration \cite{Becerik-GerberBurcin2011Aott}\cite{FranaszekMarek2009Faro}. Moreover, the targets can be placed in areas that do not affect results from the FEM modelling. 

Below, the framework for the digital target-based reconstruction of welded components is presented in further detail.

\subsection{Selection of targets}
The targets can be of various types, including spheres and high-contrast plane paper targets. In this case, circular paper stickers of equal size are chosen as target points. The placement of the targets is an important aspect, as studies have shown that non-uniformly placed targets in a line at equal height can result in inaccuracies in the computed spatial transformations \cite{Becerik-GerberBurcin2011Aott}. The chosen targets are, therefore, placed uniformly on the fixture in a semi-random pattern with a density that ensures that a minimum of 3 targets are present in two overlapping point clouds. The targets (circles) can be seen on the sample in Fig.~\ref{fig:sample}. 
	
\subsection{Pre-processing of the point cloud}
Initially, noisy data is removed from the scan to achieve a clean point cloud. This is primarily done through a statistical approach, where the Euclidean distance of each point is compared to a region of its neighbouring points. If its distance exceeds the global mean and standard deviation, the point is considered an outlier and removed \cite{Rusu2008}. 

\subsection{Alignment and merging}
The initial course alignment is based on the defined robotic scanning trajectories, while the subsequent fine adjustment is based on the approach proposed by \cite{AkcaDevrim2003Faro}. However, the proposed approach in this work makes use of captured backscatter from the laser to create a 2D image grey-scale image of the scanned surface, wherein the circular targets are identified and located based on the Hough transform \cite{YuenHk1990CsoH}. Note that a dictionary is created such that a transformation back to 3D is possible. 
The principle is essentially to pair matching targets between overlapping scans, i.e., by locating the same circles in each overlapping scan, it is possible to compute the spatial transformation between the scans. 

\begin{figure}[!t]
	\centering
	\includegraphics[width=1\linewidth,trim={0cm 0cm 0 0},clip=false]{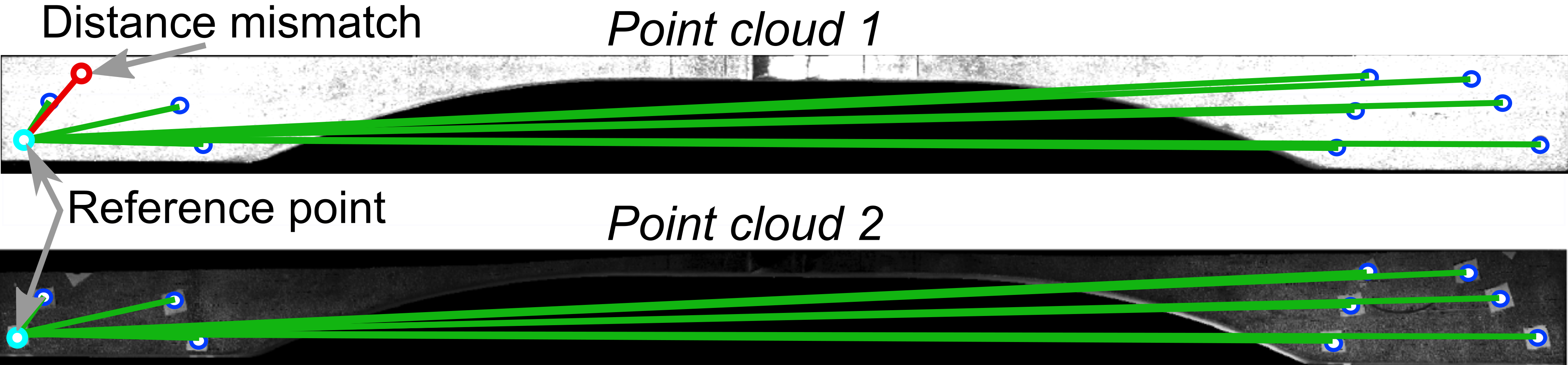}
	\caption{The principle behind filtering out targets. Point clouds 1 and point cloud 2 depict overlapping areas. The blue markers indicate the centre positions of the located targets. The green lines illustrate the matching distances between the two overlapping point clouds, point cloud 1 and point cloud 2. The red line and point illustrate a unique distance with an associated point that only exists in point cloud 1; hence, the point is omitted. Note that the example is based on a randomly selected reference point, however, the operation is carried out for all points.}
	\label{fig:filtering}
\end{figure}

\begin{figure}[!t]
	\centering
	\includegraphics[width=1\linewidth]{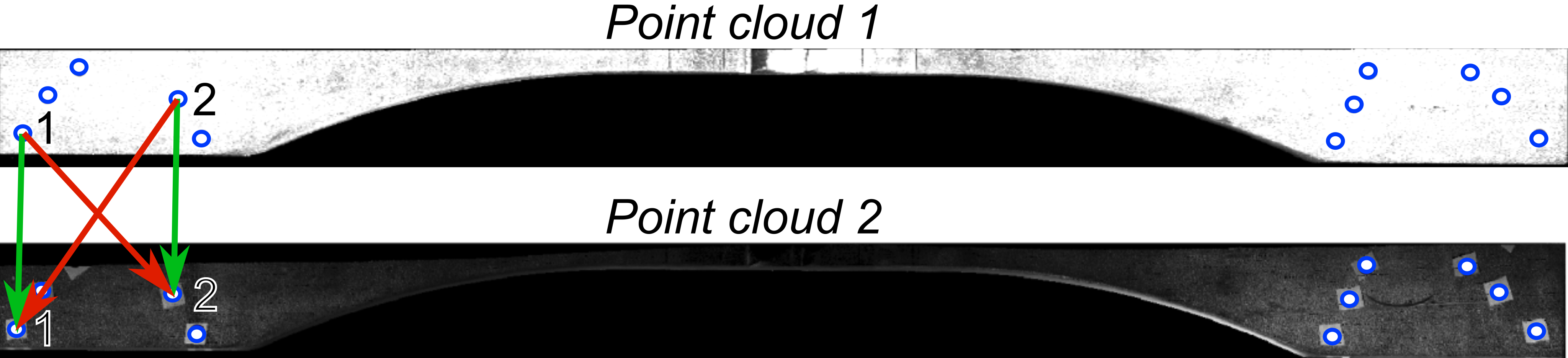}
	\caption{The principle behind matching the targets. Point clouds 1 and point cloud 2 depict overlapping areas. Note that the green lines are parallel and with an equal length for the matching points (1 and 1) and (2 and 2) in the two point clouds. Due to the assumption of pure translation, this indicates that the targets match. Instead, the crossing red lines indicate a mismatch, as such a match would require a significant rotational transformation. The example presents the use of points 1 and 2, although the operation is carried out for all points that have been selected using the principle presented in Fig. \ref{fig:filtering}.}
	\label{fig:matching}
\end{figure}

The approach consists of two main steps: 
\begin{enumerate}
    \item \textbf{Filtering outlier circles:} The first step is to filter out any target not present in overlapping point clouds. The approach is to compute the Euclidean distance between all targets in all possible combinations in each scan. Next, the list of computed distances of each scan is compared to each other. If a unique distance (within a specified tolerance) is found, and the associated target does not have other matches, it is omitted from the list of potential matches. The principle is illustrated in Fig.~\ref{fig:filtering}. 
    \item \textbf{Matching circles:} The second step is to match the targets between the scans. As the points clouds are rigid and an initial course alignment has been performed, it is possible to assume a pure translation. Therefore, circles are matched by ranking the Euclidean distances between the overlapping point clouds. The distance with the most similarities is utilised as the reference for selected matching target pairs. The principle is illustrated in Fig.~\ref{fig:matching}. It should be noted that the assumption of pure translation is only applied when matching the targets and not when computing the transformation.  
\end{enumerate}

With matched target pairs, it is possible to compute the spatial transformation of the chain of overlapping point clouds with a point-to-point approach based on \cite{Arun1987LeastSquaresFO}. As such, the transformation is determined through a non-linear optimisation, which minimises the Euclidean distance between the target pairs. After alignment, the point clouds must be merged to create a homogeneous resolution. This is done by applying a box grid filter that merges overlapping points. The entire process is performed automatically. The use of prevalent and simple image processing operations removes the dependency on expensive commercial software solutions.  

\subsection{Removal of unwanted objects}
A connectivity analysis segments the different objects in the point cloud to remove unwanted objects. Since the targets have been placed on the sample, it is possible to use this information to determine which segmented objects are part of the sample. Any unwanted objects, such as parts of the fixture, can then be removed from the scan. In other cases, the known size of the geometry could also have been used as a feature to identify it.

\subsection{Surface reconstruction}
The aligned and merged point cloud is automatically imported into Meshlab using command-line integration. Meshlab is an open-source program for processing and editing 3D meshes. In Meshlab, the normals of the point cloud are computed using neighbouring points. This is followed by a screened Poisson surface reconstruction, which meshes the final surface, creating a watertight surface. Lastly, the reconstructed representation is exported as a .stl file, which can be imported into the FEM software of choice, such as ANSYS.  

\begin{table*}[!t]
\caption{Overview of the scanning parameters and the duration of scans. Note that the resolutions refer to the local coordinate system of the scanner, while the scan direction refers to the global coordinate system of the robot. The $x$-axis points along the length of the sample, while the $y$-axis points along the width.}
\centering
    \begin{tabular*}{\linewidth}{@{\extracolsep{\fill}}lrrrrrrr}
    \toprule
    Scan direction              & Res. $x$-axis       & Res. $y$-axis    & Res. Z     & Velocity & Length of scan & Time per pass & Number of passes \\ \midrule
    Along length of sample ($x$-axis) & $\sim$0.02 $\tfrac{\si{mm}}{px}$ & 0.1 $\tfrac{\si{mm}}{px}$ & $\sim$0.004 $\tfrac{\si{mm}}{px}$ & 15 $\tfrac{\si{mm}}{s}$  & 630 $\si{mm}$ & 42 $\si{s}$          & 24          \\
    Along width of sample ($y$-axis) & $\sim$0.02 $\tfrac{\si{mm}}{px}$ & 0.1 $\tfrac{\si{mm}}{px}$ & $\sim$0.004 $\tfrac{\si{mm}}{px}$ & 15 $\tfrac{\si{mm}}{s}$  & 160 $\si{mm}$ & 11 $\si{s}$          & 8           \\ \midrule
    Total                  & ~            & ~         & ~             & ~        & ~      & $\sim$18 min      & 32          \\ \bottomrule
    \end{tabular*}
    \label{tab:scanSpec}
\end{table*}

\section{Experimental setup}\label{sec:exp}
\begin{figure}[!t]
	\centering
	\includegraphics[width=0.9\linewidth]{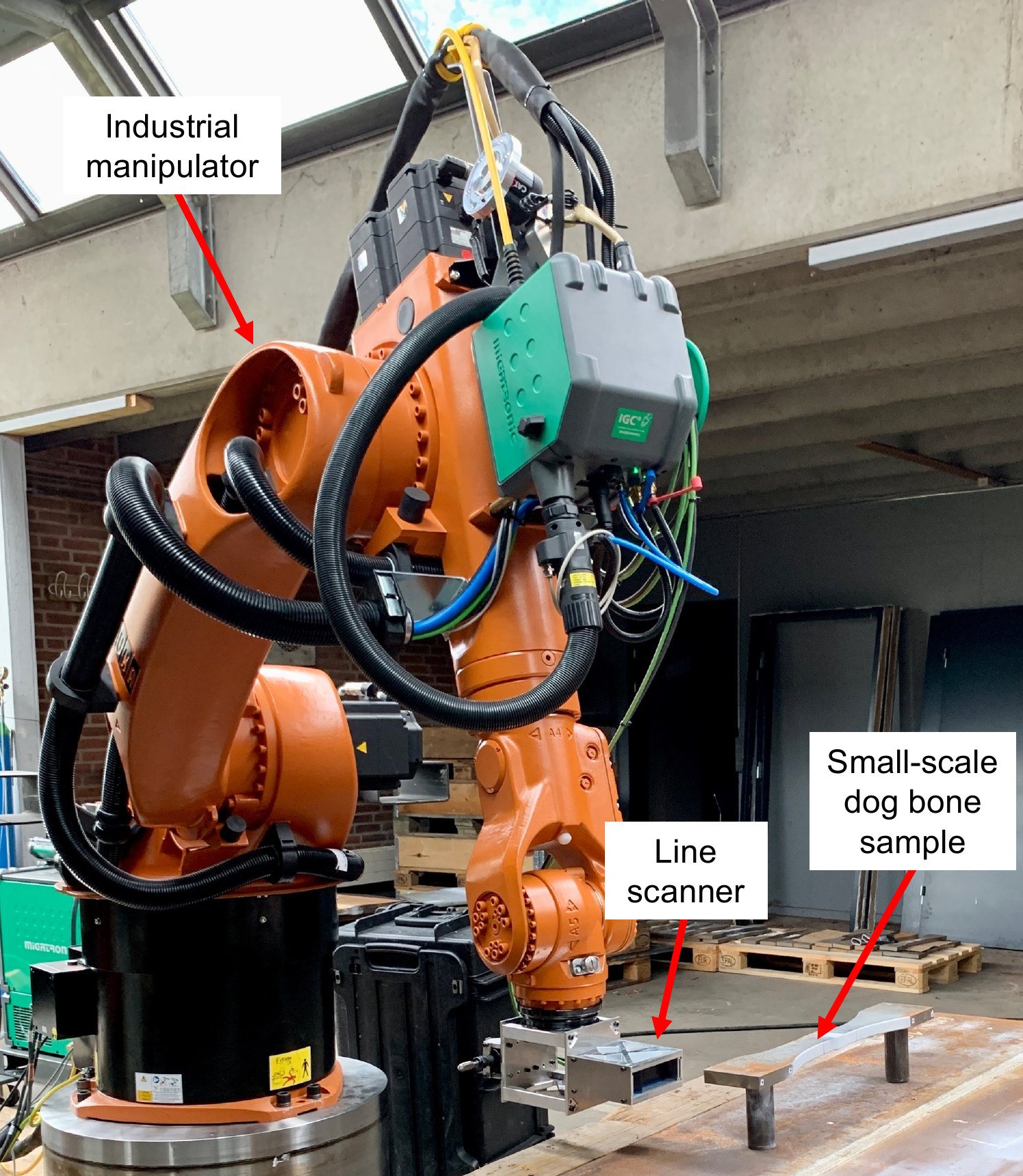}
	\caption{The experimental setup, consisting primarily of the industrial manipulator and the line scanner.}
	\label{fig:expSetup}
\end{figure}

The experimental setup is similar to the setup presented in \cite{Mikkelstrup2022}, designed to perform automated post-weld treatment. The setup, illustrated in Fig.~\ref{fig:expSetup}, consists of two main components, a KUKA KR60-3 industrial manipulator and a Wenglor MLWL131 line scanner. All computations are performed in MATLAB 2019b and Meshlab v.2020.07 with an Intel Core i7 (i7- 9750H) CPU @ 2.6 GHz.

As stated in Section \ref{sec:method}, the testing sample is a small-scale sample of an offshore jacket structure which is welded using gas metal arc welding in a double-sided joint configuration. The sample is made of S355 construction steel and has been water jet cut into the shape of a dog bone with a size of 100 $\times$ 570 $\times$ 20 $\si{mm}^3$ for fatigue testing.

All robotic trajectories have been manually programmed and follow a linear trajectory in which the velocity and orientation are kept constant throughout the entire motion. Further details of scanning parameters can be found in Table \ref{tab:scanSpec}.

Before scanning, the sample must be prepared. This includes removing any surface contamination, such as paint and rust, around the weld area to avoid any geometrical details being concealed. Additionally, the sample is sprayed with scanning spray to reduce reflectivity (when scanning reflective samples), and the reference targets (stickers) are manually and semi-randomly placed for alignment. Upon scanning, the sample must be roughly positioned at a specific position ($\pm$ 10 $\si{mm}$) to ensure it is correctly scanned and merged.

\section{Results and discussion}\label{sec:res}
\begin{figure*}[!t]
	\centering
	\includegraphics[width=1\linewidth]{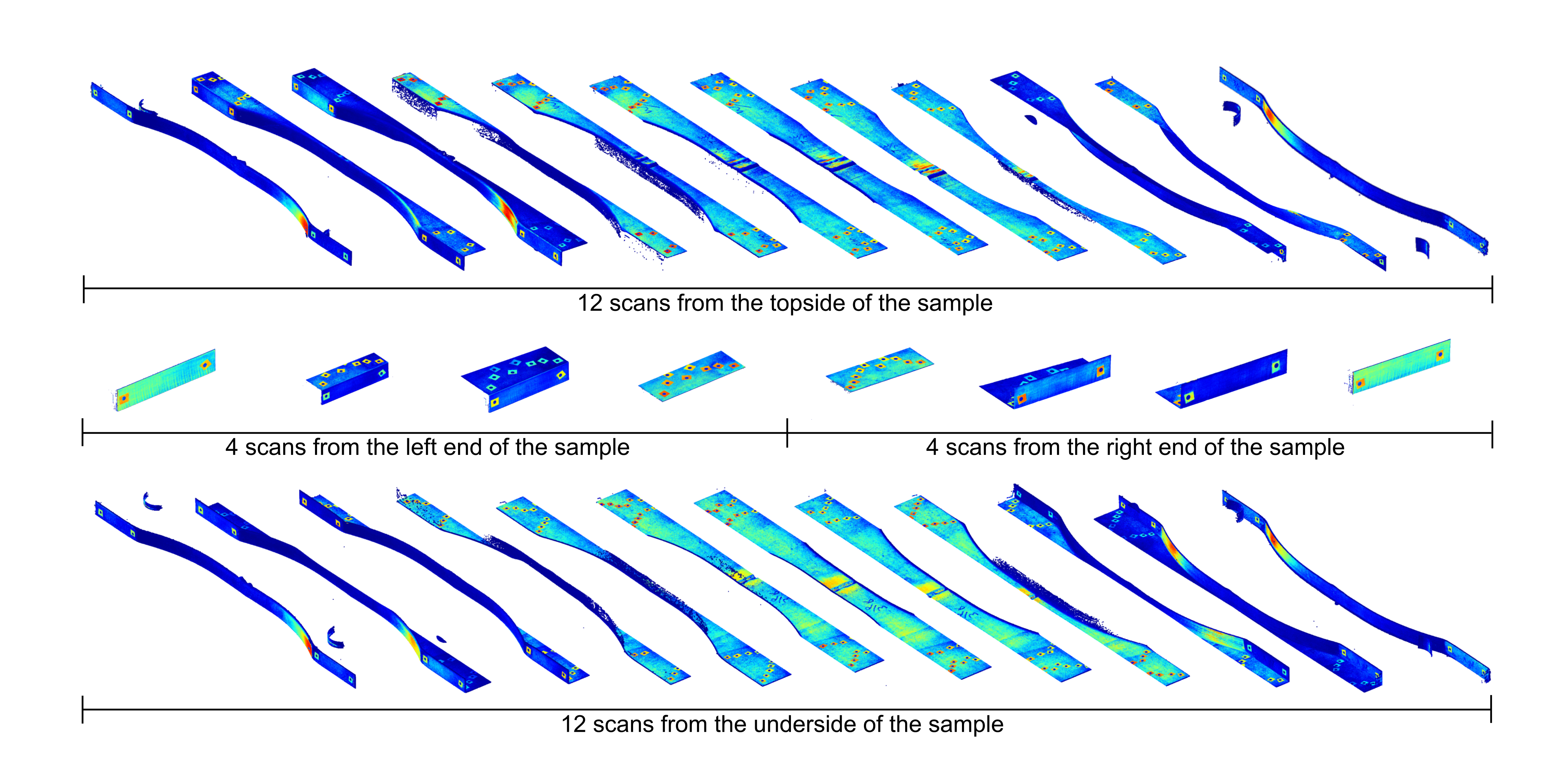}
	\caption{Depiction of all 32 scans required to form a complete representation of the scanned samples.}
	\label{fig:combinationOfScans}
\end{figure*}

Due to the line scanner's limited field of view (FoV), 32 scanning passes are performed to ensure adequate overlay between individual scans, depicted in Fig. \ref{fig:combinationOfScans}. To cover the bottom half of the sample, it is manually rotated 180\degree~around a single axis. In total, the scanning duration of a sample is $\sim$30 min, where $\sim$18 min is pure scanning time. This is clear from   Table \ref{tab:scanSpec}. Note that with the applied scanner, the scanning speed and the resolution between the scan lines ($y$-direction) are co-dependent. Hence, the scanning speed can be increased at the expense of the resolution. Similarly, the FoV can be expanded by moving further away from the sample, but it results in a lower resolution. Lastly, the scanning time can be reduced by optimising the overlay, which has not been done in this work. 

\begin{figure*}[!t]
	\centering
	\includegraphics[width=0.8\linewidth]{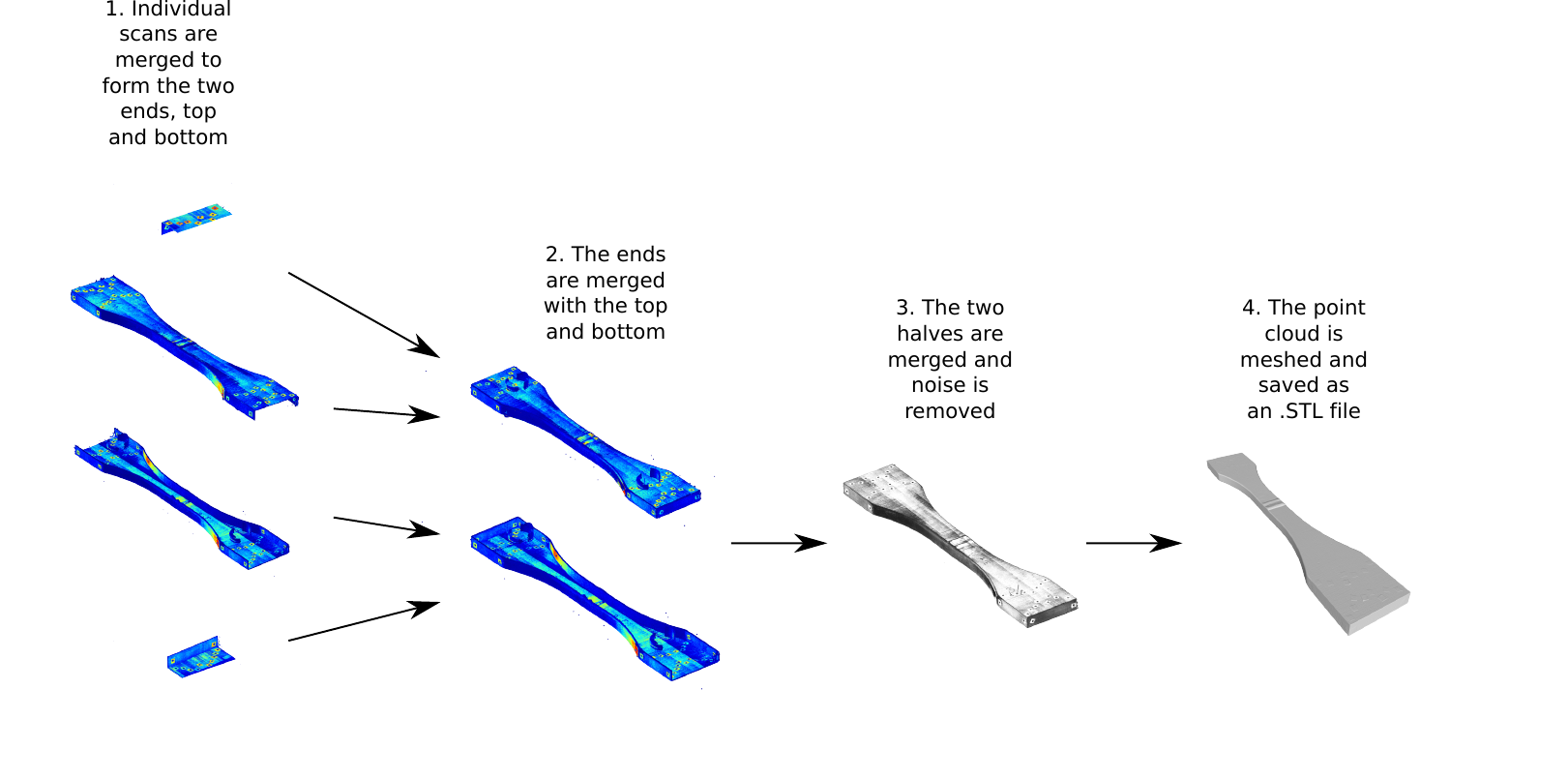}
	\caption{The operations required to reconstruct the chosen sample. First, the 32 individual scans of the ends, top and bottom, are merged and aligned to form the two halves, which are then merged, and the noise is removed. Lastly, the sample is meshed, reconstructed, and saved as a .stl file for FEM modelling. Note the detail included in the representation, such as the reference targets and the course edges due to water jet cutting.}
	\label{fig:stepsForRe}
\end{figure*}

Having scanned the sample, the individual point clouds are then aligned, merged and denoised based on the principles presented in Section \ref{sec:method}. This resulted in a complete point cloud of $\sim$34,000,000 points, represented as a $\sim$900 MB ASCII file with a resolution of 0.1 $\tfrac{\si{mm}}{px}$. The ASCII file is then     imported into Meshlab, where the normals are computed using 100 neighbouring points. This is followed by the screened Poisson surface reconstruction, where a reconstruction depth of 12 has been chosen to ensure a high level of detail. The computational time is $\sim$15 min; however, the implementation is not optimised. The result is a 504 MB .stl file consisting of $\sim$5,000,000 vertices and $\sim$10,000,000 facets. Reducing the construction depth could reduce the file size, but also reduces the level of detail. The procedure is illustrated in Fig.~\ref{fig:stepsForRe} along with the final reconstructed result.

\begin{figure}[!t]
	\centering
	\includegraphics[width=1\linewidth]{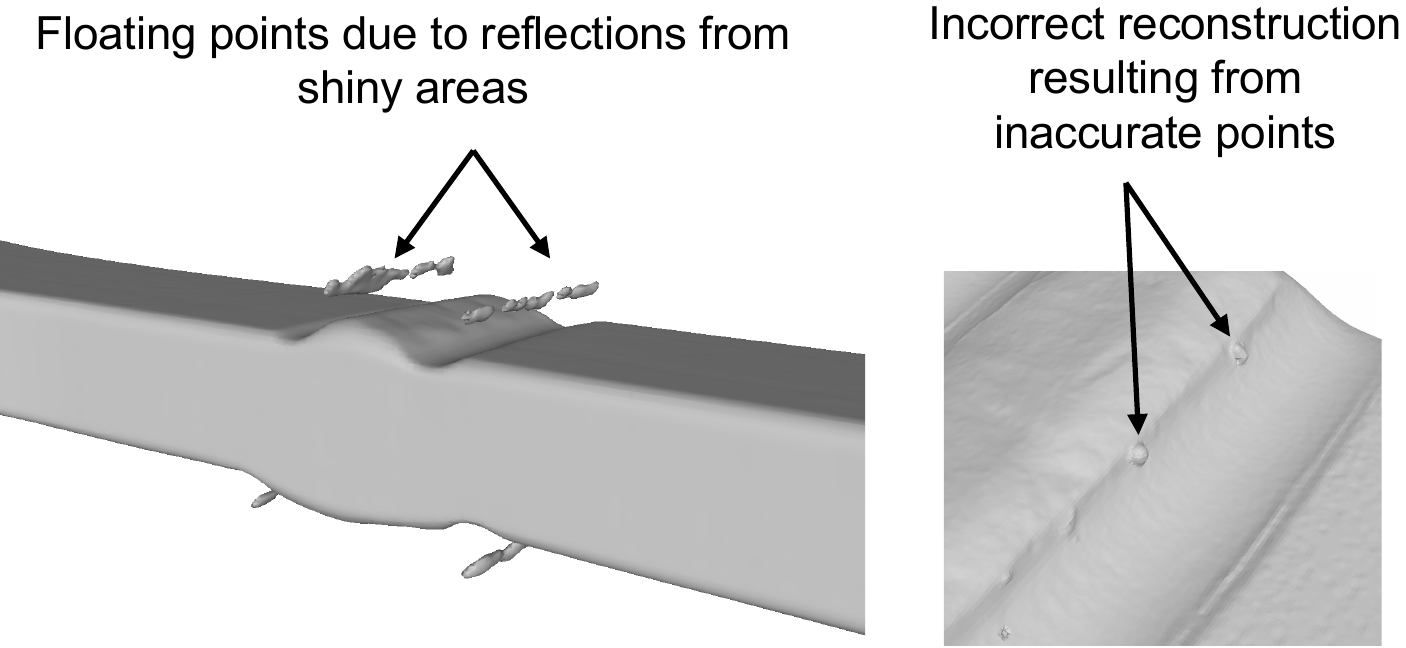}
	\caption{Observed errors in reconstructed samples. The illustration depicts a burr-grinded sample, representing a case where the treated areas are highly reflective.}
	\label{fig:burrGrind}
\end{figure}

When closely observing the reconstructed sample, slight misalignment of overlapping scans can be noticed ($<$0.05 $\si{mm}$). This is due to several factors, such as the scanning speed or orientation not being constant, but primarily due to incorrect detection of the target centres. Other errors have also been observed, such as floating points due to spurious reflection or false representation of the surface from noisy points. Even a pen ink on the surface affects the reconstruction, underlining the sensitivity of the process. The problems primarily occur when scanning highly reflective surfaces. This is illustrated in Fig.~\ref{fig:burrGrind}, which depicts the errors observed when reconstructing a treated sample. Since the sample is scanned from all directions, the lack of visible misalignment indicates that the framework provides an accurate representation. However, to determine the precise accuracy of the approach, the proposed framework must be applied to evaluate the dimensions of a reference part with known dimensions.

The results demonstrate that the framework effectively reconstructs the welded component with sufficient accuracy for fatigue life prediction through FEM modelling.  

\section{Conclusion}\label{sec:conc}
The proposed framework allows for the digital reconstruction of welded samples directly in the production setup using an industrial manipulator and a line scanner. The framework relies on prevalent image processing techniques and non-linear optimisation to align and merge several point clouds, representing the entire geometry of the sample. This is followed by a screened Poisson surface reconstruction to form the final 3D representation. Focus has been placed on developing a generic approach that does not rely on specific hardware or expensive commercial software. The result is a flexible, convenient, and cost-effective approach that can be applied for a range of uses, including predicting the fatigue life improvements of the treatment, aiding at the design stage of the component based on knowledge of the possible crack location, and improved quality assurance and documentation of automated HFMI treatment.

The future work includes some aspects, including optimisation of the scanning procedure, e.g., by using robotic simulation software to reduce the number of scans and targets and, thereby, the processing time. Similarly, a non-uniform point cloud resolution could be applied, such that the resolution is increased near the weld where a high level of detail is required and decreased away from the weld. This would result in a reduced file size and reduced computational time. Lastly, rotating the sample could be performed with a rotational axis controlled by the robot to avoid manual interference. 

\section*{Acknowledgements}
The authors would like to gratefully acknowledge Joachim Emil Kokholm Hersbøll for providing insight into FEM modelling of welded components. 

\bibliographystyle{IEEEtran}
\bibliography{IEEEabrv,mybib}

\end{document}